# Benchmarking local Hebbian learning rules for memory storage and prototype extraction

Anders Lansner[1,2,*], Andreas Knoblauch[3], Naresh B Ravichandran[1], Pawel Herman[1,4]

[1]KTH Royal Institute of Technology, EECS, Stockholm
[2]Stockholm University, Dept. of Mathematics, Stockholm
[3]KEIM-Institute, Albstadt-Sigmaringen University, Germany
[4]Digital Futures, KTH Royal Institute of Technology, Stockholm

*Corresponding author: ala@kth.se

**Abstract**

Associative memory or content-addressable memory is an important component function in computer science and information processing, and at the same time a key concept in cognitive and computational brain science. Many different neural network architectures and learning rules have been proposed to model the brain's associative memory while investigating key component functions like figure-ground segmentation, perceptual reconstruction and rivalry. A less investigated but equally important capability of associative memory is prototype extraction where the training set comprises distorted prototype instances and the task is to recall the correct generating prototype given a new distorted instance. In this paper we benchmark associative memory function of seven different Hebbian learning rules employed in non-modular and modular recurrent networks with winner-take-all dynamics operating on moderately sparse binary patterns. We measure pattern storage and weight information capacity, prototype extraction capabilities, and sensitivity to correlations in data. The original additive Hebb rule comes out with worst capacity, covariance learning proves to be robust but with moderate capacity, and the Bayesian-Hebbian learning rules show highest capacity in almost all different conditions tested.

## Introduction

Associative memory as a concept in computer science refers to a memory that is content addressable, i.e., able to retrieve a stored item when given a fragment or distorted copy of it or to retrieve an item when cued by another associated item. Such so-called auto- and hetero-association, respectively, also reflects the meaning of associative memory in cognitive brain science and psychology. Associative memory capabilities of recurrent cortical neural networks are thought to underlie fundamental pattern processing operations like e.g. figure-ground segmentation, perceptual completion and rivalry, as well as long-term memory Click or tap here to enter text.. Key elements of stimulus-response behavior and associative chaining may be described as heteroassociative, with a stimulus item associated to, e.g., an action or a successor in a word sequence.

The search for the neural mechanisms underlying human associative memory dates back at least to Donald Hebb's cell assembly theory and hypotheses about mental representations in the form of cell assemblies and memory based on synaptic associative, i.e. "Hebbian", plasticity (Hebb, 1949). Fundamental questions about memory engrams and the mechanisms behind our brain's cognitive and perceptual functions have been pursued in experimental brain research for long time and are still on top of the agenda (Josselyn & Tonegawa, 2020).

The focus of this paper is on functional aspects of one-layer autoassociative memory networks and related neuroscientific theories and computational models. Following Palm (2013), the term "neural associative memory" (NAM) is adopted to distinguish such neurally oriented models of associative memory from non-network models often studied in cognitive science and from other common types of error-correction based artificial neural networks. Given the neurobiological plausibility of Hebb's hypothesis about associative learning in the brain, we focus here on different variants of Hebbian learning rules proposed for associative memory. Our aim is to compare quantitatively by means of a set of benchmark tasks the associative memory capacity of these different learning rules when employed in a single layer, recurrently connected NAM network model. To emphasize and better reflect the influence of the learning rules employed, we aim to use generic procedures with minimal network architectures and activation functions.

Recent extensions to the basic NAM include mechanisms to generate hidden layers and higher order internal representations, for example predictive coding models (Tang et al., 2023), modern Hopfield networks (Krotov & Hopfield, 2021), sparse quantal Hopfield networks (Alonso & Krichmar, 2024), and Bayesian-Hebbian learning with structural plasticity (Ravichandran et al., 2025). These more complex models demonstrate significantly enhanced capabilities compared to the classical Hopfield network (Hopfield, 1982). However, since they rely heavily on multi-layer architecture and mechanism beyond the single recurrently connected layer of neural units and Hebbian plasticity, they are beyond the scope of the basic learning rule comparison presented here.

### *Associative memory, pattern reconstruction and prototype extraction*

An autoassociative memory is content addressable in the sense that when stimulated with some input pattern the most similar, in some metric, among the stored patterns is recalled. In the linear "matrix memories" the patterns are simply vectors with real valued components. In the non-linear models, the patterns are binary or bipolar with components in {0,1} or {-1,1} respectively or with continuous-valued activation in the corresponding interval, e.g. [0,1]. Neuron spiking frequency when subject to sensory input may be considered as a confidence of the key stimulus being present (Meyniel et al., 2015). The pattern format that comes closest to this view is one with components in some interval between zero and maximal firing frequency, normalized to [0,1]. It is thus somewhat surprising that much work in the NAM field has used and still use a bipolar pattern activation function, possibly due to the strong influence from spin glass physics, as noted by Palm[1].

Proper function of the associative pattern processing in neural networks requires that their memory is not overloaded. When too many patterns are stored in a fixed size network, memory function gradually breaks down and recalled patterns become unstable while instead distorted and spurious memories are recalled. Central questions in the field have been and are still what learning rule and activation function gives the highest memory capacity and scaling to large network sizes.

---

[1]Palm, 2013: "… probably due to the misleading symmetry assumption (symmetry with respect to sign change) that was imported from spin-glass physics. This prevented the use of binary {0, 1} activity values and the corresponding Hebb rule and the discovery of sparseness."

Another interesting but less studied capability of NAMs is that of prototype extraction. It emerges in a basic form when training an associative memory network with a number of pattern instances generated from each of a set of prototype patterns by adding some form of distortion. When tested with new such pattern instances, the memory's ability to recall the most similar prototype pattern, which itself was never presented to the network, is quantified. Such behavior is closely related to concept and category formation in human cognition and also to clustering in data science when the number of clusters is not given. Work on such learning in classical Artificial Neural Networks (ANN) has been scarce, but see e.g. (Amari, 1977) and recent modeling of such operations in ANN (McAlister et al., 2024; Ross et al., 2017; Tamosiunaite et al., 2022)

## *Related work on Neural Associative Memory*

Hebb's work and publications in the late 1940's inspired research in early theoretical and computational brain science as well as in engineering. One focus was on recurrent spiking neural network models and testing for emergence of Hebbian cell assemblies in biological tissue. Early computer simulations by Rochester et. al (1956) failed to show formation of cell assemblies with sustained activity in a recurrently connected network of 69 spiking integrate-and-fire model neurons. However, more recent simulation of spiking recurrent neural network models with reasonable biophysical detail and Hebbian learning have demonstrated emergence of Hebbian cell assemblies manifesting as robust attractor dynamics and associative memory capabilities (D. J. Amit & Brunel, 1997; Brunel, 2000; Kopsick et al., 2024; Lansner, 1986; Roudi & Latham, 2007; Treves & Rolls, 1991). Other early more theoretical work concered associative memory in hippocampus (Marr, 1971) and concept formation in neocortex (see e.g. (Amari, 1977, 1989; Anderson et al., 1977; Nakano, 1972)

In the electronics and computer science domain the earliest associative memory work was by Steinbuch (1961). His LernMatrix was a binary or real valued crossbar associative network that took binary or normalized real valued vectors as input and produced a binary or real valued weight matrix during learning that was then used to generate output from new input. An important focus was hardware realization and several devices were produced and even used in applications (Steinbuch & Piske, 1963). Kohonen developed further the Correlation matrix memory, quite related to the LernMatrix with real valued weights (Kohonen, 1972). Associative memory also originated early in the research community around holographic associative memories (Gabor, 1968; Longuet-Higgins, 1968). Work by Willshaw et al. (1969) developed the concept of associative memory models with a binary weight matrix similar to the LernMatrix and it was followed by in depth analyses of the storage capacity and recall mechanisms of such NAMs (Knoblauch, 2010, 2011, 2016; Knoblauch et al., 2010; Knoblauch & Palm, 2020; Knoblauch & Sommer, 2016; Palm, 1980, 2013; Schwenker et al., 1996).

The interest among theoretical physicists in brain modeling and associative memory in the form of attractor neural networks was spawned by the work of Little (1974) and later popularized by Hopfield (1982), who brought into focus the analogy between spin-glass physics and brain neurodynamics. This work has been further extended and elaborated by many researchers (see e.g. (D. Amit et al., 1987; Betteti et al., 2025; Clark & Abbott, 2024; Kanter & Sompolinsky, 1986)) contributing extensive analysis of the dynamics of such systems with the occurrence of fixed-point, line- and chaotic attractors, and phase transitions in focus.

The majority of this theoretical work was based on additive Hebbian learning. The Bayesian Confidence Propagation Neural Network (BCPNN) was first introduced in the late 1980's (Lansner & Ekeberg, 1987, 1989) and later developed with a modular architecture of hypercolumns and minicolumns (Johansson & Lansner, 2007; Lansner & Holst, 1996a; Sandberg et al., 2002). The Bayesian Confidence Propagation learning rule (BCP) was derived from Bayes rule assuming independent inputs ("naïve Bayes") and is non-linear and harder to analyze than additive learning. It has been used extensively to model cortical associative memory in non-spiking and spiking forms (Chrysantidis et al., 2022; Fiebig et al., 2020; Fiebig & Lansner, 2017; Lansner et al., 2013; Lundqvist et al., 2010a, 2011). The BCP learning rule is closely related to Bayes Optimal Learning (Knoblauch, 2011) and its modular structure is similar to Potts neural network with "multi-state neurons" introduced in the late 1980's (Kanter, 1988) as well as to block coded associative memories (Gripon & Berrou, 2011; Knoblauch & Palm, 2020). Interest in this kind of modular neural network architectures has further recently risen in the context of quantum computing (Fiorelli et al., 2022).

## *Network architectures and local Hebbian learning*

Associative memory network models traditionally have a simple architecture often with just one recurrent layer. In this study, we used such a one-layer architecture with binary {0, 1} units operating on moderately sparse distributed activity patterns. Our focus on sparse distributed patterns is partly motivated by data on the estimated activity distribution (Tang et al., 2019) and activity levels of neurons in mammalian neocortex based on energy calculations and single unit recordings. These findings indicate that typically less than 1 % of pyramidal cells are active in neocortex at any instant (Lennie, 2003; Quiroga, 2012; Waydo et al., 2006). In our study, such sparse activity input patterns are considered as having originated from a sensor array, possibly feeding into a layer of composite feature detectors, or some equivalent input source.

Two types of network configurations were considered in our study: a non-modular one with $N$ units and a modular one having $H$ modules ("hypercolumns") each comprising $M$ units ("minicolumns). Minicolumns represent smaller local vertical clusters of some hundred more densely connected neurons observed in primate neocortex (Kaas, 2013; Mountcastle, 1997; Opris & Casanova, 2014; Powell et al., 2024; Rockel et al., 1980; Wallace et al., 2022). For this kind of modular networks, the partitioning of the $N$ network units could be done in many ways, but we here followed the "small world" scheme, i.e. $H = M = \sqrt{N}$ proposed for cortex by Braitenberg (1978). For network sizes with above some hundred modules, increase is expected to take place by expanding the number of hypercolumns.

The state update equations for the $j^{th}$ neural units was the standard:

$$h_j = b_j + \sum_i^N x_i w_{ij}$$

where $N$ is number of units, $h_j$ is the field, $b_j$ is a bias, $x_i$ is the presynaptic unit activity and $w_{ij}$ is the weight of the connection from presynaptic unit $i$ to postsynaptic unit $j$. For the non-modular network we used k-winners-take-all (kWTA) whereas in the modular network, each module used a local WTA (single winner) activation function. For comparison of non-modular and modular types of networks, $K$ was chosen equal to $H$, resulting in the same number of active units. Other more capable and biologically plausible activation functions have been proposed and evaluated, but for our purpose here of comparing different learning rules we judged the simplest ones to be most appropriate.

We employed iterative updating during memory recall, making the recurrent network operate as a so called attractor associative memory (Amit, 1990). Such a retrieval scheme has been shown to enhance storage capacity (Schwenker et al. 1996). See the Methods section for more information.

In our comparison we considered only local learning rules of a Hebbian type, employing correlation based learning via simple probabilistic measures of neuron activity and co-activity available at the synapse (Minai, 1997; Stuchlik, 2014; Zenke et al., 2015). We also included learning rules that feature intrinsic plasticity, i.e. an activity dependent regulation of unit baseline activity, as has been described experimentally (Egorov et al., 2002) We compared the following seven selected learning rules with acronyms as follows (see also Table 2):

1. The WILL learning rule proposed by Willshaw (1969) based on earlier work by Steinbuch (1961) and later analyzed extensively by Palm et al., see e.g. (Palm, 2013);
2. The HEBB learning rule, see e.g. (Amari, 1977);
3. The sparse HOPF learning rule based on (Hopfield, 1982) for binary neural units, later adapted for sparse activity patterns ((Amari, 1989);
4. The COV learning rule proposed by Tsodyks and Feigelman (1988), adding a term to the Hopfield learning rule, making independently active units develop zero weights between them;
5. The PRCOV learning rule proposed by Minai (1997) with the intent to improve storage capacity for correlated patterns;
6. The BOM learning rule derived from probabilistic Bayesian considerations, here used without dependence on input or output noise, called BOMs, (Knoblauch, 2011), see Appendix A.
7. The BCP learning rule also derived from probabilistic Bayesian considerations proposed by Lansner and Ekeberg (1989) and later adapted to a modular neural network architecture (Johansson & Lansner, 2007; Lansner & Holst, 1996);

In all cases, training was conducted sample-by-sample and in one-shot mode, i.e. each pattern in the training set was presented once during training, and learning was based on a simple frequentist approach, as detailed in Tables 1 and 2. In the case of non-modular networks the self-connections (autapses) were set to zero. For modular networks the connections between units in the same hypercolumn were set to zero, leaving WTA normalization (Carandini et al., 1997) as their only interaction.

| Counter variables | p-estimates |
|---|---|
| $C = \sum_k 1$ | - |
| $C_i = \sum_k x_i^{(k)}$ | $p_i = C_i/C$ |
| $C_j = \sum_k x_j^{(k)}$ | $p_j = C_j/C$ |
| $C_{ij} = \sum_k x_i^{(k)} x_j^{(k)}$ | $p_{ij} = C_{ij}/C$ |

**Table 1.** Equations used to accumulate the amount of unit activations and co-activations during training in counter variables, from which $p$-estimates were calculated for derivation of weights and biases in Table 2. Here $k$ indexes patterns, $x_i$ and $x_j$ are binary training pattern components in {0, 1}, where $i$ and $j$ correspond to indexes of pre- and postsynaptic neural units.

The Hebbian learning rules included here can be properly expressed in terms of activity and co-activity statistics (Minai, 1997), as formulated as in Table 2.

| Abbrev. | Bias | Weight |
|---|---|---|
| WILL | - | $1 \; if \; p_{ij} > 0, \;$ 0 otherwise |
| HEBB | - | $p_{ij}$ |
| HOPF | - | $p_{ij} - a[p_i + p_j] + a^2$ |
| COV | - | $p_{ij} - p_i p_j$ |
| PRCOV | - | $\frac{p_{ij} - p_i p_j}{p_i}$ |
| BOM | $(n-1)\log\frac{1-p_j}{p_j} + \sum_{i=1}^{n} \log\frac{p_j - p_{ij}}{1 - p_i - p_j + p_{ij}}$ | $\log\frac{p_{ij} \cdot (1 - p_i - p_j + p_{ij})}{(p_i - p_{ij}) \cdot (p_j - p_{ij})}$ |
| BCP | $\log p_j$ | $\log\frac{p_{ij}}{p_i p_j}$ |

**Table 2.** Equations for computing bias and weight values for different learning rules expressed in terms of activity and co-activity statistics and formulated as probabilistic $p$-estimates (cf. e.g. Minai, 1997) as in Table 1. Here $a$ is activity density, i.e. fraction of ones in a pattern. We numerically stabilized all fractions by imposing a lower bound $\varepsilon$ on all denominators and, within $\log(.)$, also on numerators (see Methods section).

Figure 1 illustrates that different learning rules results in quite different weight trajectories and final weights range during learning of the same set of patterns. Some learning rules produce negative weights. We assume this negative component represent some kind of disynapic inhibitory plasticity in real cortex (Chrysanthidis et al., 2019).

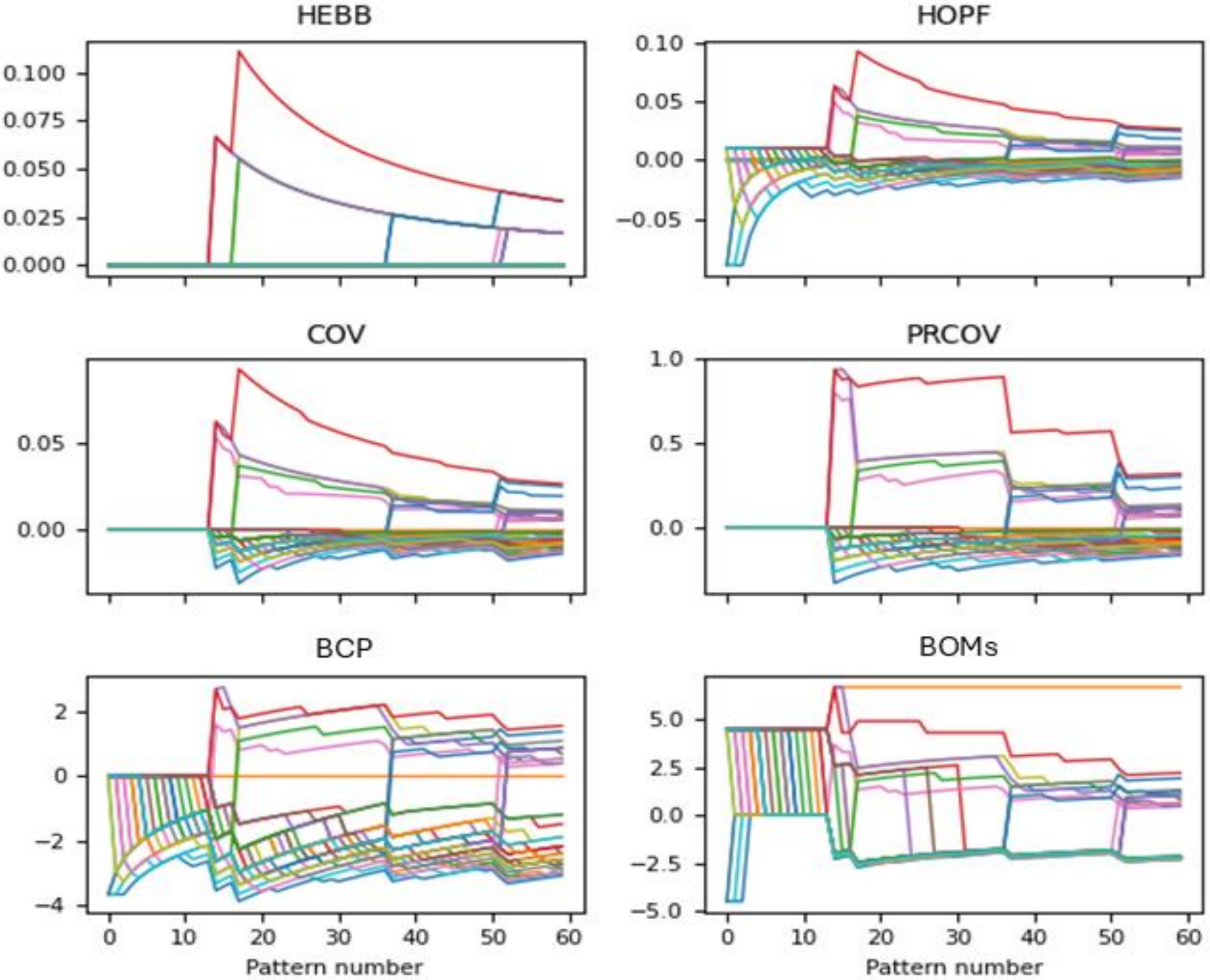

**Figure 1: Weight trajectories from different learning rules.** Different learning rules produce quite dissimilar weight trajectories when trained on the exact same set of 60 training patterns. The network was modular with 10 hypercolumns, each comprising 10 neural units. The values of the same set of 40 weights are shown. The WILL learning rule with binary weights was excluded.

## *Experimental setup and evaluation*

The evaluation of the different learning rules was based on the network performance in four main tests:

1. *Pattern storage capacity* i.e. given by the maximum number of stored patterns at which 90% could be recalled error-free from test patterns (cues) distorted by input noise. The number 90% is somewhat arbitrary but was used consistently in the following.
2. *Weight information capacity (C)* , i.e. total stored information in bits divided by number of trainable weight,
3. *Prototype extraction*, according to same criterion as in 1, but recalling prototypes rather than individual patterns. Training and test instances were generated by adding input noise to the generating prototype, and

4. *Correlation resistance,* by monitoring the decline in pattern storage capacity with increasing amount of correlation among pattern components in the training set.

In cases 1 and 3 we investigated how the performance scaled with growing network size, whereas for weight information a saturating value with size was obtained. All test were performed for non-modular as well as modular network configurations.

Types of sparse random patterns

Two types of sparse random patterns with subtypes depending on network architecture were used:

1. Standard random binary patterns. For non-modular networks $K$ active units were activated randomly and for modular networks, one random unit per hypercolumn was set active.
2. Correlated random patterns. These were generated by the method described by Minai (1997) giving the correlation parameter $f_p$ some different values. This introduced broadened distributions of unit usage and also violationed the "naïve Bayes" assumption of independent pattern components.

For modular networks, the input noise (pattern distortion) for creating test patterns as well as training instances from pattern prototypes were introduced by random resampling of a fraction (typically 10, 25, and 50 %) of hypercolumns in a pattern. For non-integer values of distortion the floor() and ceil() values were mixed proportionally to achieve a proper mean fraction. For non-modular networks the unit activations themselves were resampled in the same way.

Memory capacity and recall

The left panel in Figure 2 exemplifies how the fraction of error-free recall ($f_{corr}$) depends on number of patterns in the training set. As the network load increases the recall fraction declines crossing the dotted line at P90. This is used as a measure of the pattern capacity of the different learning rules. Further details can be found in the Methods section.

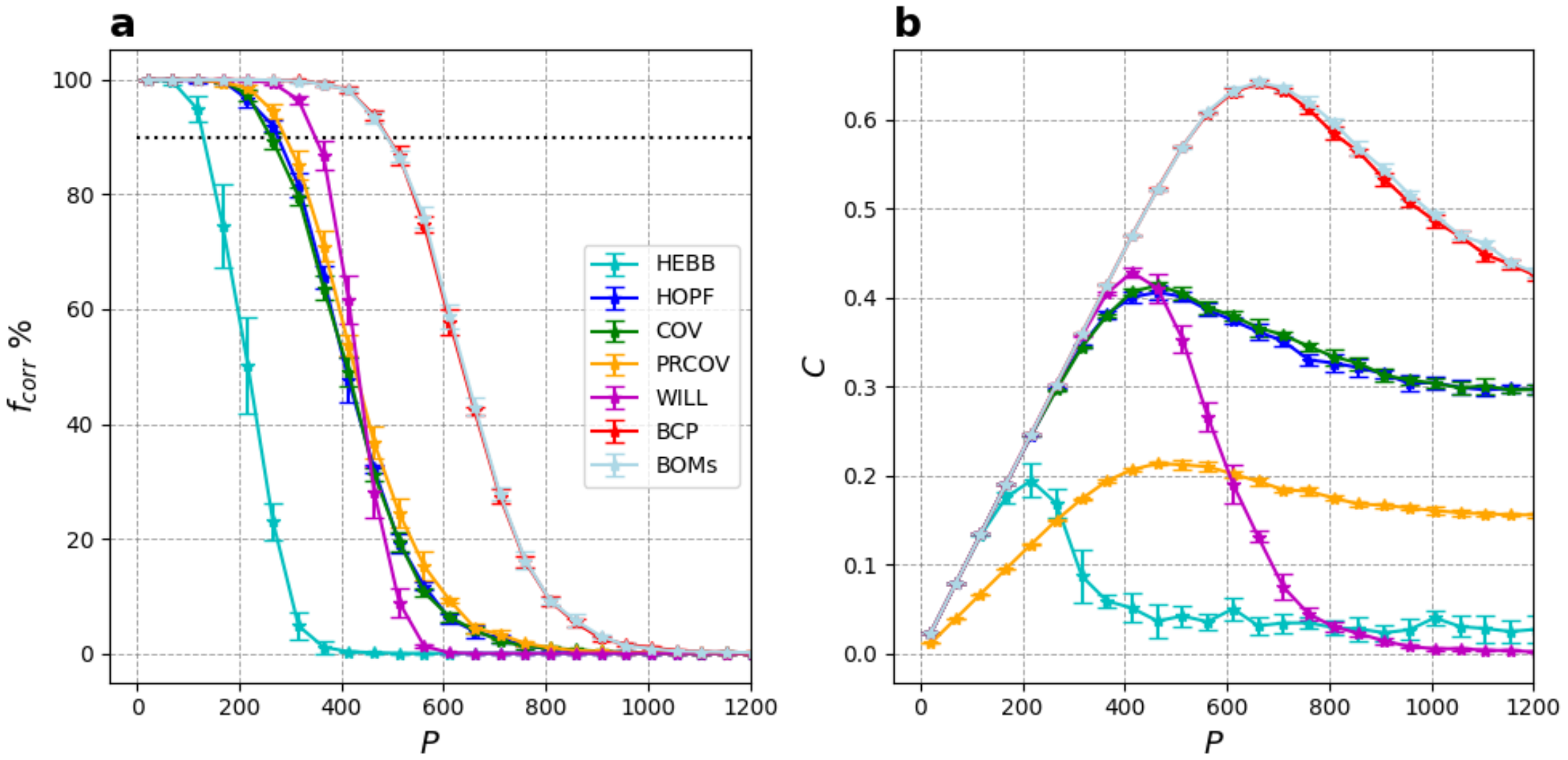

**Figure 2: Recall fraction and weight information of a modular network.** Dependency on number of patterns ($P$) in the training set and different learning rules. The network had 20 hypercolumns each with 20 units and was trained with increasing number of patterns. Test patterns had 10% of hypercolumns resampled and error bars show standard deviation. (a) Recall fraction fell to low values with increasing memory load. The dotted line marks 90% error-free recall. (b) Weight information capacity (*C*) showed an inverted-U shape with a maximum of 0.64 at $f_{corr}$= 50%. The low values of PRCOV is due to its asymmetric weight matrix. Each data point is based on 20 runs with different pseudo random seeds per data point.

However, the memory capacity measured like this is also highly dependent on network size, network architecture, and activity density ("sparsity") of the patterns, or more precisely, the information content of each pattern. Patterns with few active units contain less information and can be stored reliably in higher numbers than more dense patterns. A complementary storage capacity measure in this context is weight information capacity (*C*), i.e. the number of bits of information stored in the network relative to the number of free parameters (trainable weights). This measure is more independent of network size, architecture, and activity sparsity and therefore suitable for comparison of different learning rules. It was quantified as described in the Methods section. The right panel in Figure 2 shows how *C* changes with increasing number of stored patterns. Notably, the maximum for each rule is obtained at $f_{corr}$ around 50%. This maximum is below one bit and well below the ca 5 bits a biological synapse can represent (Bartol et al., 2015). The observations were qualitatively the same for a trained non-modular network.

Prototype extraction and recall

For prototype extraction, several random prototype patterns were created and from each of them a number of distorted instances was generated. Distorted patterns were used both for training and testing. The task of the network was to recall the generating prototype pattern from a distorted version of it (Figure 3). As can be seen from the bottom row, the network was able to reconstruct one of the prototypes almost perfectly when trained with ten or more instances and tested with new unseen instances. Notably, the calculation of the mean in the middle row was done given information of from which prototype a pattern instance was generated. This information was not given to the network, which therefore solved a harder problem.

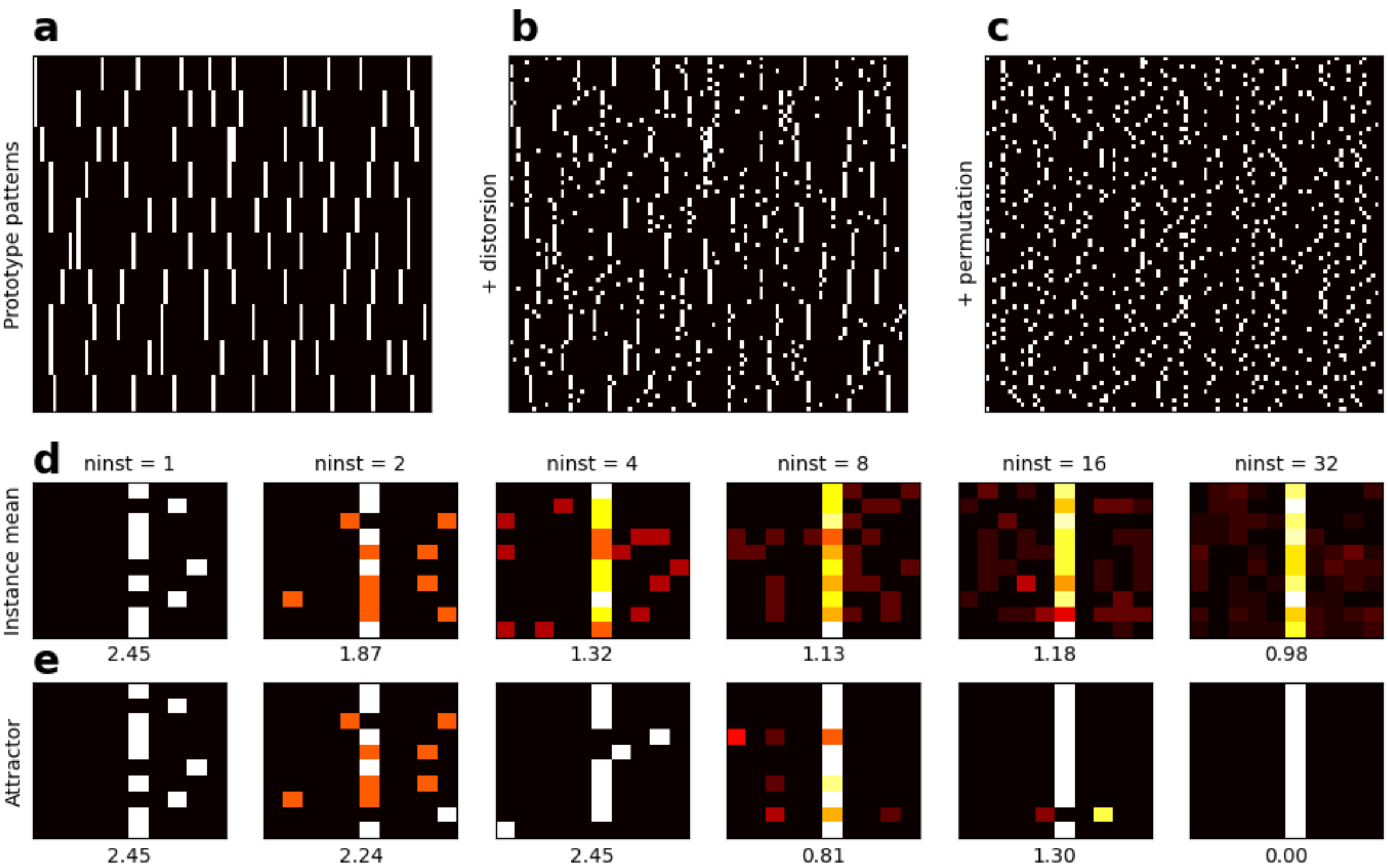

**Figure 3: Prototype extraction from different number of instances. (a)** 10 random $H = 10$, $M = 10$ prototype patterns, one per row. Please note that the pattern on the last row, if unfolded in 2D, forms a vertical bar as seen in panels **d** and **e**. **(b)** Shows 10 training instances from each of the 10 prototypes. Each instance was resampled in 3 randomly selected hypercolumns. **(c)** The training instances in randomly permuted order forming the final training set. **(d)** Mean of the instances of the last prototype pattern unfolded in 2D for increasing number of instances per prototype (ninst). **(e)** The stable attractor state unfolded in 2D reached by a $H$=10, $M$=10 BCPNN trained with the 'ninst' instance patterns and tested with new instance patterns. The number below each panel in **d** and **e** gives the average Euclidean distance between attractor state reached given the last prototype as input.

## Results

In this section we give quantitative results on pattern storage capacity, weight information capacity, and prototype extraction for the seven different learning rules selected, for different network architectures. This is done with randomly generated binary patterns and at different levels of input noise. In addition we test the capacity of the different learning rules to maintain performance despite correlations in patterns. Results are mainly derived in the form of how the number of correctly recalled patterns or prototypes scale with network size.

### *Pattern capacity scaling*

Figure 4 shows how the pattern capacity at the 90% error-free recall level scaled with network size for non-modular (upper panels) and modular (lower panels) networks. It was evident that the storage capacity varied with learning rule and that some of these were more sensitive than others to input noise. The HEBB rule showed lowest storage capacity among all whereas HOPF, COV and PRCOV had intermediate performance and BOMs and BCP rules were best performing. The former was slightly ahead at low and intermediate input noise while the latter performed best at high input noises.

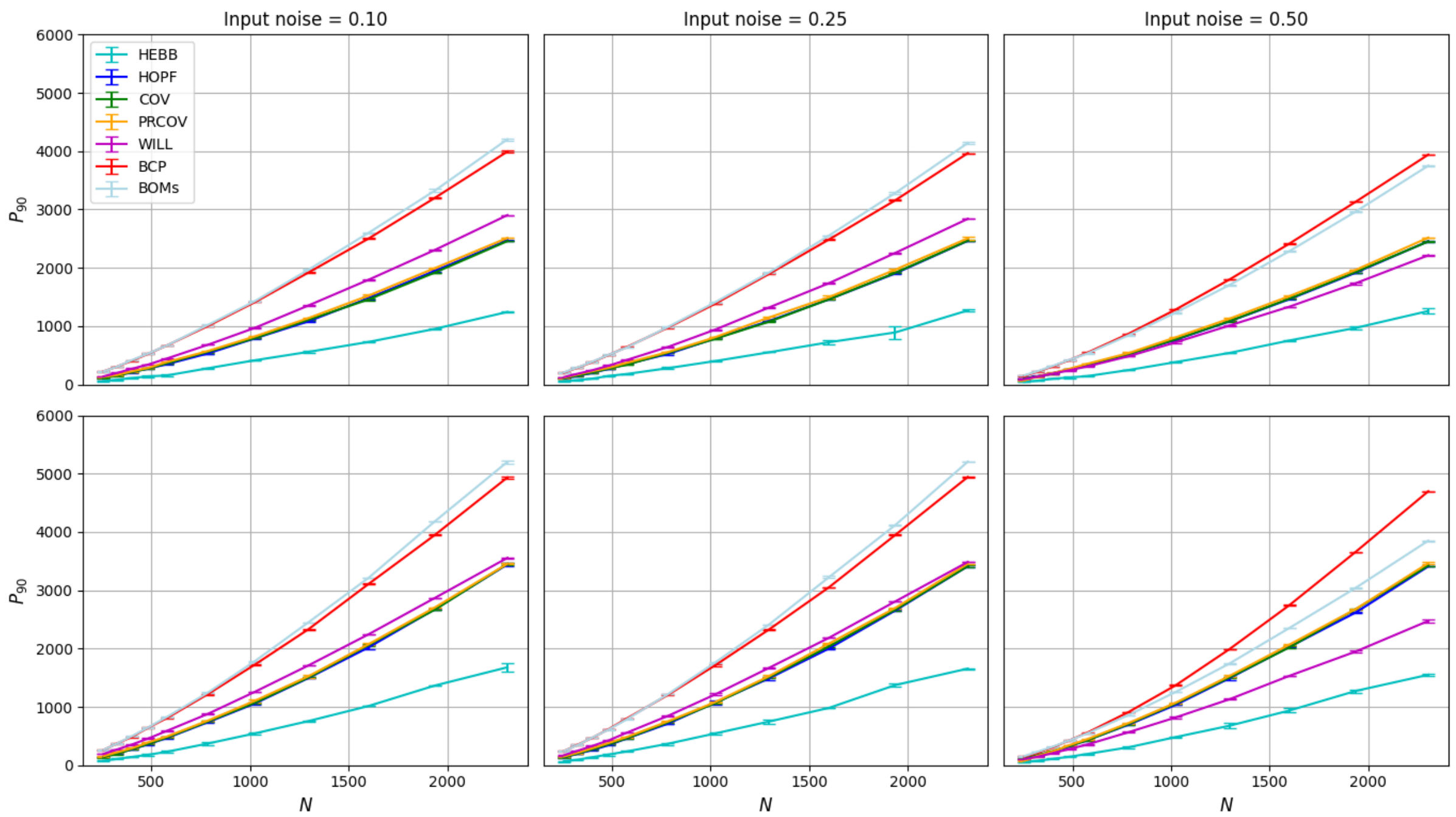

**Figure 4. Pattern capacity of non-modular and modular networks.** Pattern capacity depending on input noise and learning rule, measured as the maximum number of patterns that allows for an error-free recall fraction of 90% ($P_{90}$). The upper row corresponds to non-modular networks and the lower to modular networks. Each data point is the mean and standard deviation of 5 runs. The legend in the upper left panel holds for all panels.

## *Weight information capacity*

Weight information capacity (*C*) in bits was measured in the same runs that provided data for Figure 4. We measured *C* to be comparable to previously published values on bit per weight often given in theoretical work (see Methods). Figure 5 gives means over the largest network sizes showing that learning rules can also be ranked according to *C*-values. We can observe much the same trends as for pattern capacity when it comes to the rank and sensitivity to input noise of the learning rules. A difference is that the PRCOV rule has lower rank due to the asymmetric weight matrix it produces, which reduces *C* to half. Table 3 gives the numeric values at input noise level 0.10. We are here in the "robust retrieval regime" so these values should be comparable to previously reported theoretical *C* values (see Methods). We note that network architecture has negligible if any effect on *C*.

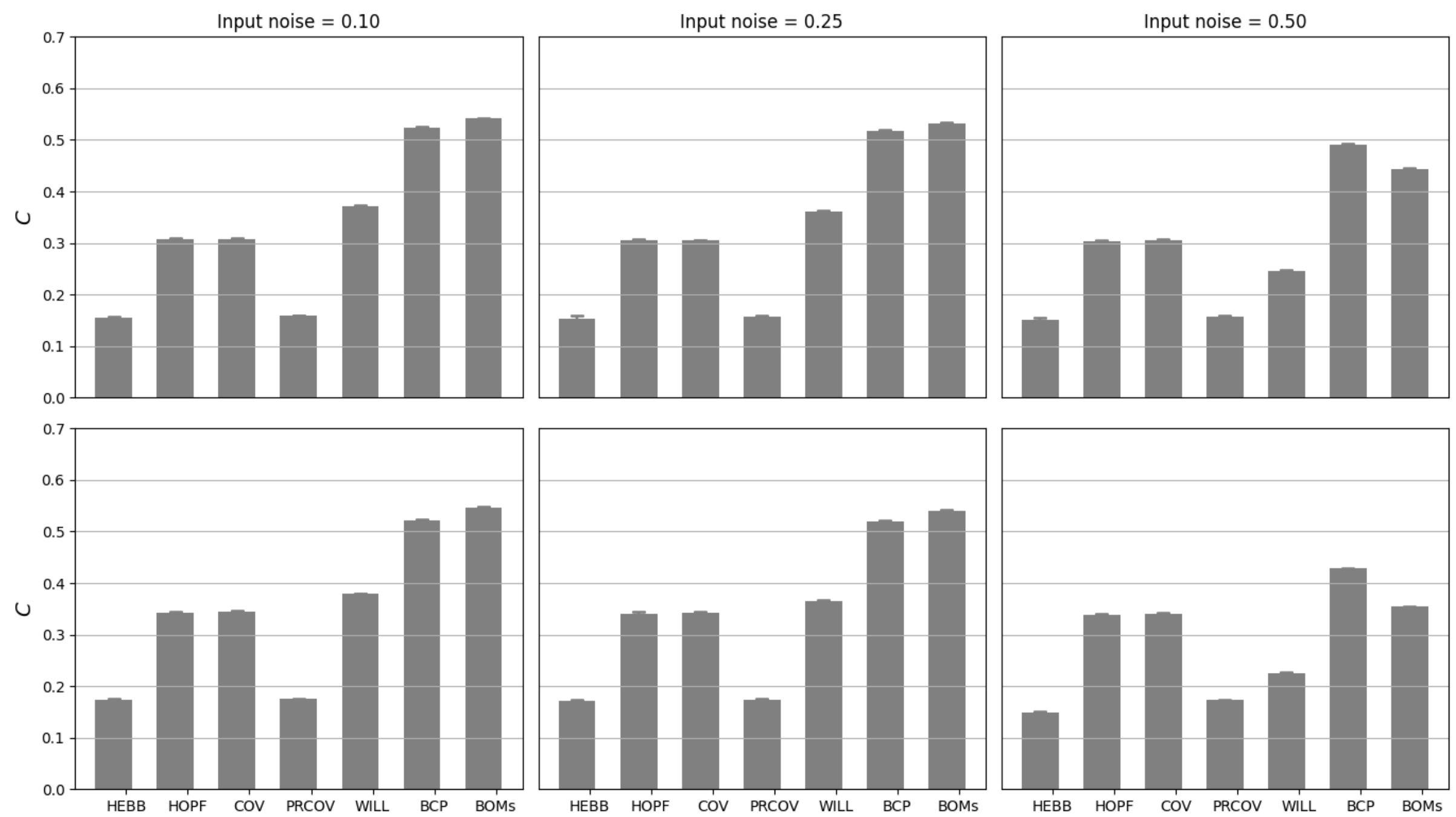

**Figure 5. Weight information capacity of non-modular and modular networks.** Mean values of weight capacity ($C$) for networks of size $N$ = 2304 with $K/H$ = 48, depending on learning rule and level of input noise at recall. Non-modular networks in upper and modular ones in lower row. The low values of PRCOV is mainly due to its asymmetric weight matrix. Each data point is the mean and standard deviation of 5 runs.

| | HEBB | HOPF | COV | PRCOV | WILL | BCP | BOMs |
|---|---|---|---|---|---|---|---|
| Non-modular | 0.16 | 0.31 | 0.31 | 0.16 | 0.37 | 0.52 | 0.54 |
| Modular | 0.17 | 0.34 | 0.34 | 0.17 | 0.38 | 0.52 | 0.55 |

**Table 3.** $C$ in bits per weight at input noise level 0.10 for non-modular and modular networks.

## *Comparison to earlier theoretical and empirical results*

We have been serching the literature for $C$-values to compare with for the case of moderately sparse activity. In fact, there is a lack of such values, in particular from simulations. However, we note that the value for WILL (0.37) is very close to the theoretically derived ln(2)/2 (0.35) in the sparse limit given already by Palm (1980).

An often used relation given for the dependence of pattern storage capacity on activity density was given by Tsodyks and Feigelman (1988) as

$$\alpha_c = \frac{P_c}{N} = \frac{C_{TF}}{a|\ln a|} \qquad (1)$$

where $\alpha_c$ is the critical capacity, $P_c$ is critical pattern capacity, $a$ is activity density of patterns stored and $C_{TF}$ is a parameter, typically 0.1 – 0.2.

We can adapt this relation for our case were $a = 1/\sqrt{N}$ which gives

$$P_c(N) = \frac{2C_{TF}N^{2/3}}{\log N} \qquad (2)$$

Figure 6a shows the theoretical relationship in eq. 1 together with some simulated data from the literature and from the present study. Here as well, data is surprisingly scarce and when given obtained from quite variable conditions with regard to, for instance, capacity measure used, feed-forward or recurrent network architecture, iterative retrieval or not, and type of patterns and noise used. This makes quantitative comparison difficult and inconclusive at this stage.

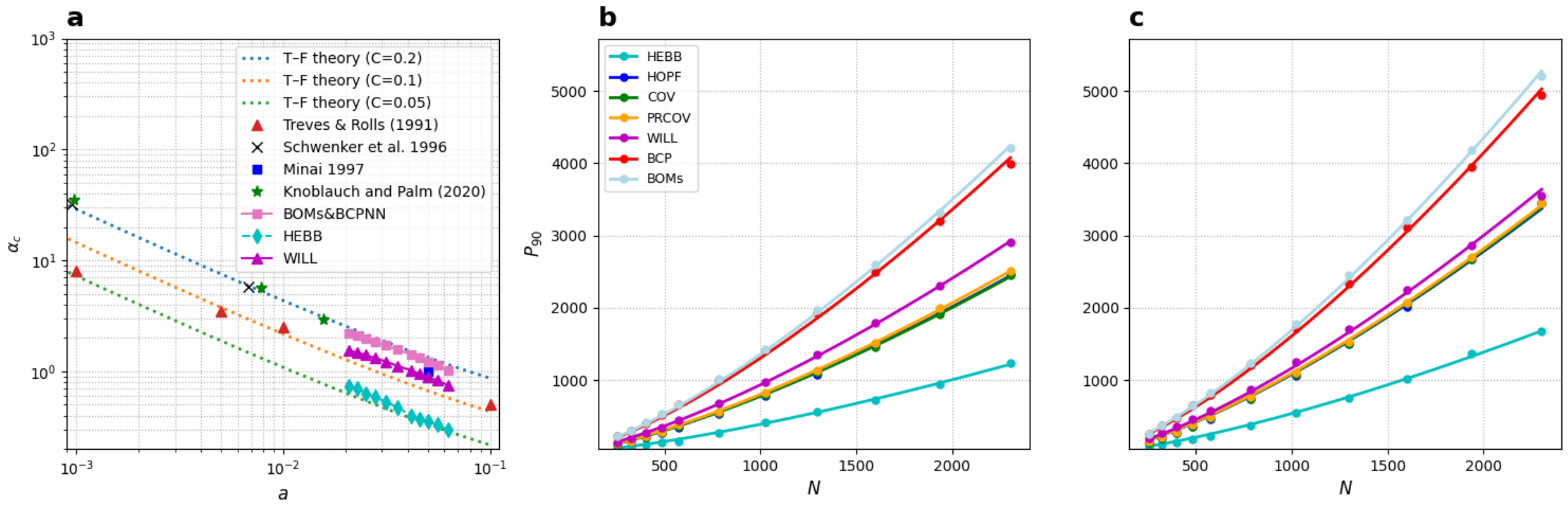

**Figure 6. Fit with theory and earlier simulation results. (a)** Plot of dependency of $\alpha_c$ on pattern activity density ($a$) for modular networks. The dotted lines show the theoretical relations given by eq. 1 for parameter $C_{TF}$ in $\{0.05, 0.1, 0.2\}$. Included are also a number of published values obtained from the sources in the legend. **(b)** Fit of eq. 2 to pattern capacity data points in Figure 4 at input noise level 0.1, for non-modular network format. **(c)** same as (b) but for modular networks.

Figure 6b, c show that our empirical pattern capacity data points from Figure 4 are well fitted by the theoretical relation of Tsodyks and Feigelman and thus scale with network size as predicted from eq 2, giving each of our seven learning rules a numerical value for the parameter $C_{TF}$ (Table 4). The rankings are the same as for the pattern capacity and also for $C$ except for PRCOV.

| | HEBB | HOPF | COV | PRCOV | WILL | BCP | BOMs |
|---|---|---|---|---|---|---|---|
| Non-modular | 0.043 | 0.086 | 0.086 | 0.088 | 0.102 | 0.143 | 0.148 |
| Modular | 0.059 | 0.118 | 0.119 | 0.120 | 0.127 | 0.176 | 0.184 |

**Table 4.** Value of parameter $C_{TF}$ for the fits in Figure 6.

## *Prototype extraction*

The prototype extraction capabilities of different network architectures and learning rules were evaluated in a similar manner as for storage of individual patterns. The main difference was that instead of training with a number of individual patterns, it was done with instances generated by distortion from a set of prototypes. The networks were tested for ability to reconstruct the generating prototype from unseen distorted test instances.

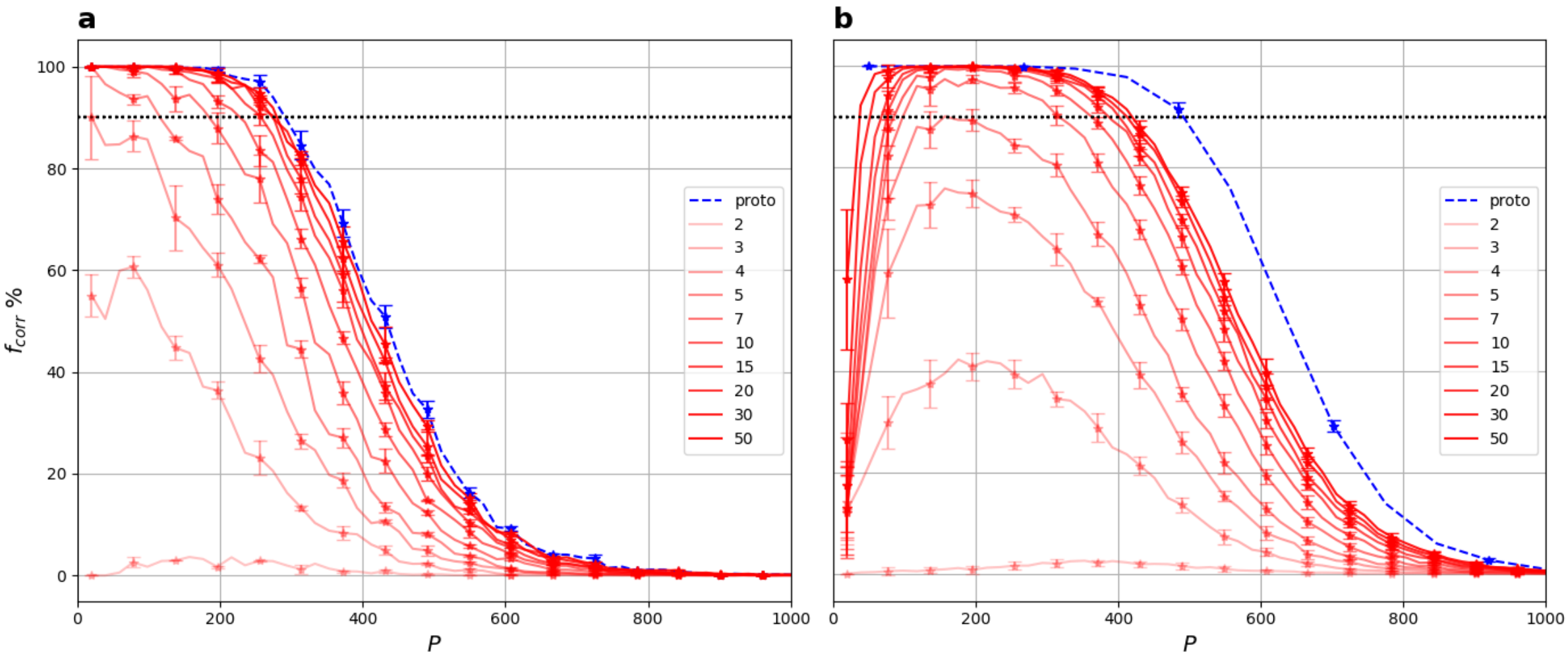

**Figure 7: Fraction of correct recall of prototypes.** Pattern capacity of a modular *H* = 20 *M* = 20 network trained with variable number of prototype patterns and number of instances (see legend) generated from those, at an input noise level of 10%. The task was to recall the generating prototype without error. The dashed blue line shows performance when only the prototype patterns themselves were stored. The dotted line marks the 90 % recall fraction and error bars show standard deviation from 10 runs. The learning rule in **(a)** was PRCOV and in **(b)** BCP.

We found a qualitative difference between the Bayesian-Hebbian learning rules and the other five in the memory low load region. Figure 7 illustrates this difference in how pattern storage capacity for prototype patterns for two representative learning rules, PRCOV and BCP, depended on the number of prototypes and instances per prototype in the training set. As can be seen, both learning rules allow good prototype extration performance. Notably, it takes only about five instances for error-free prototype recall to reach the 90% level. But they differ also in two important respect. For low number of prototypes (*P*), PRCOV quickly builds a prototype and recalls it when new instances are presented, whereas BCP is slower at doing the same. Furthermore, when trained with a high number of instances the PRCOV pattern capacity approaches the same performance as with the prototypes themselves in the training set (Figure 7, dashed curves), whereas BCP performace stayes well below that, though still higher than for PRCOV. Further simulations showed that the qualitative behavior of PRCOV was shared with HEBB, HOPF, and COV learning rules and BOMs performed similar to BCP. We also observed that these Bayesian-Hebbian learning rules had a higher tendency to generate attractors for individual training instances at low *P* and also to produce spurious attractors close to the prototype state. This resulted in low performance because individual instances and spurious states were recalled instead of the prototypes. The generalization performance of different Hebbian learning rules at low memory load in such prototype extraction tasks obviously require a deeper investigation and detailed comparison to how animals behave in corresponding situations. This even relates to generalization and identification of outliers in human concept formation (Fernandino et al., 2022). Although quite interesting this is beyond the scope of the current work.

The scaling of pattern storage capacity with network size could be measured in the same way as for individual patterns, by extracting $P_{90}$ at the crossing with the dotted line at 90% $f_{corr}$. Figure 8 shows results for the two network architectures, the seven learning rules and different pattern input noise. As can be seen from the overall lower performance, the task was generally harder than when storing individual patterns (compare to Figure 4). It is not surprising that the WILL learning rule failed entirely due to the fact that a single pre/postsynaptic coincidence is sufficient to switch the synapse from 0 to 1. The Bayesian-Hebbian rules (BCP and BOMs) again showed highest pattern storage capacity with a good margin.

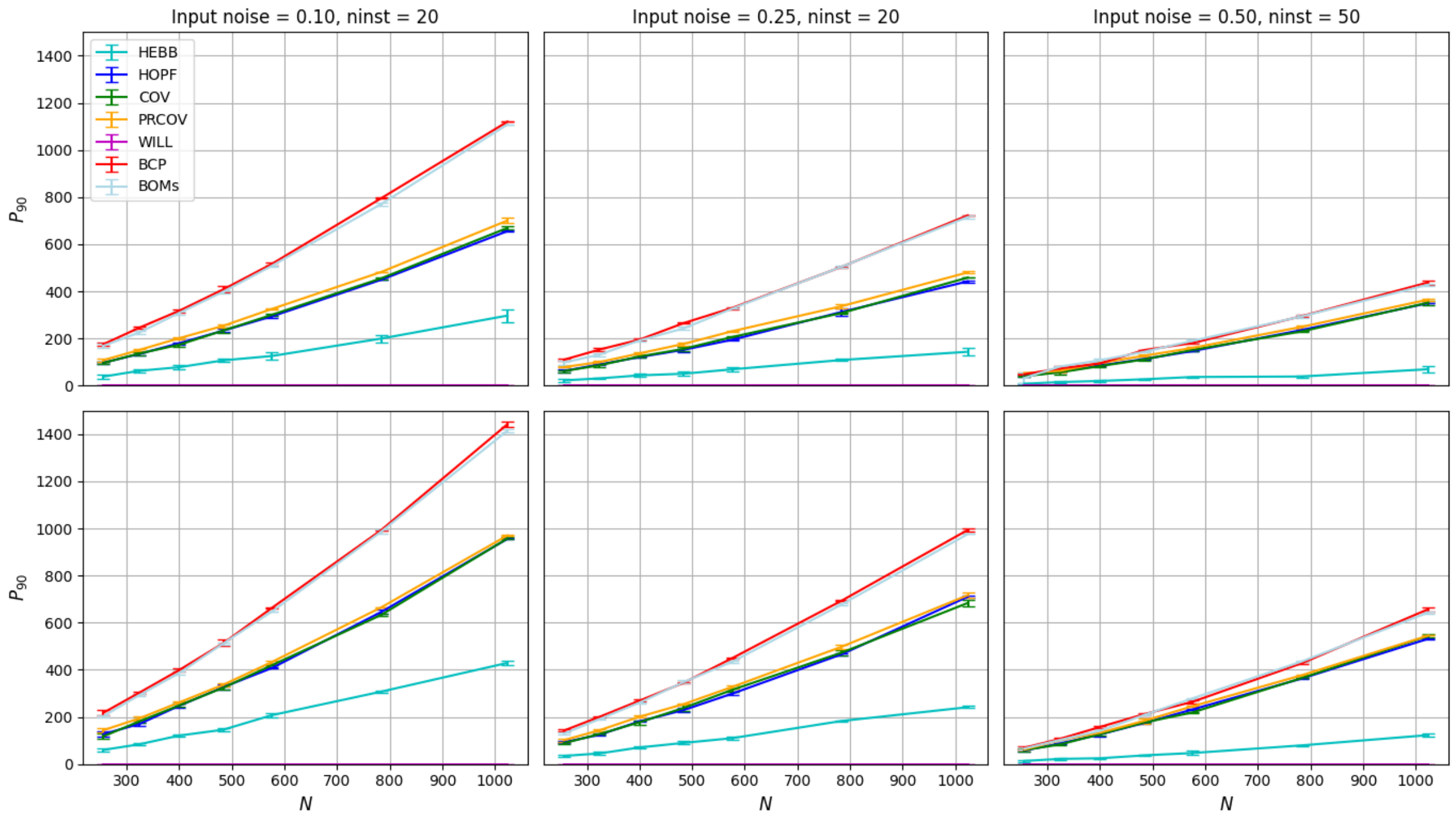

**Figure 8: Prototype storage capacity scaling.** Performance measured as the maximal number of prototypes allowing error-free prototype recall at 90% depending on input noise level and learning rule. Upper row refers to non-modular networks, lower row to modular networks. The number of training instances generated from each prototype was 20 or 50, higher for a noise level of 0.50. Each data point is the mean and standard deviation of 5 runs. Legend in upper row left panel holds for all panels.

## *Sensitivity to correlated data*

We have so far only considered standard sparse random binary patterns. Real pattern datasets typically contain correlations and one-layer associative memory networks are known to show reduced storage capacity for such datasets. The presynaptic covariance learning rule (Minai, 1997) was specifically designed to achieve a higher robustness to correlations in data.

Figure 9 shows how pattern storage capacity of different learning rules is affected by training and testing with correlated data, for non-modular and modular networks when storing individual patterns as well as during prototype extraction. From the obtained datapoints a “correlation resistance” index was calculated and used in the final summary table (Table 5). Preliminary test indicated that results were invariant to network size .

The rank order of different learning rules varied somewhat but consistently the WILL and HEBB rules showed highest sensitivity. The PRCOV learning rule was consistently among the least sensitive, especially in the prototype extraction task. The Bayesian-Hebbian learning rules were equally good as PRCOV for single pattern storage but fell back significantly in the prototype extraction task, in particular for non-modular networks. For higher correlations (*fP* values) PRCOV showed better pattern capacity even in absolute terms than the Bayesian-Hebbian learning rules.

A related study and results on associative memory correlation sensitivity can be found in Minai (1997). Ours is the first to include Bayesian-Hebbian learning rules and it could be extended further, but a more in depth investigation is left for future study.

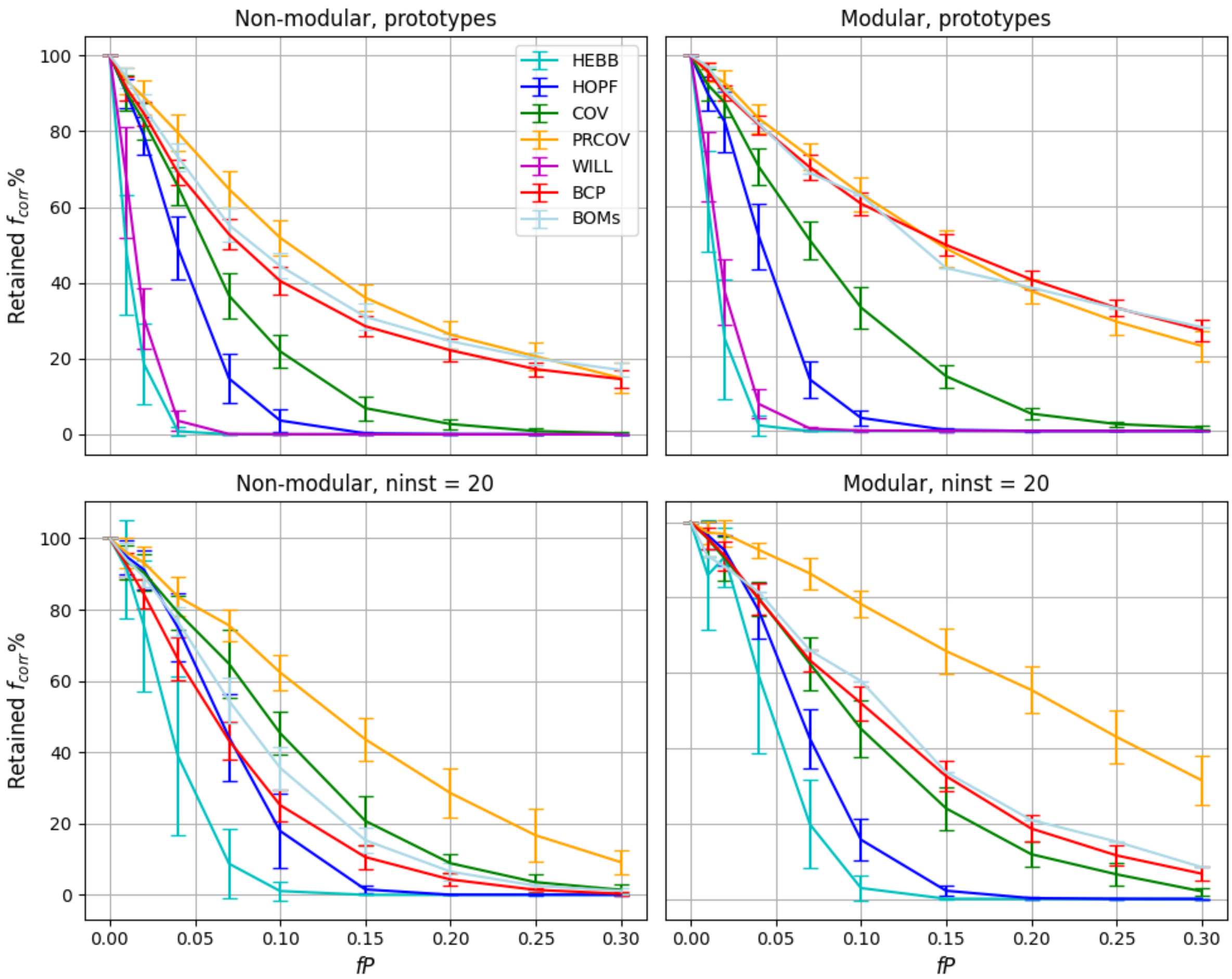

**Figure 9. Sensitivity to correlations in data of non-modular and modular networks.** The figure shows the ratio between $fcorr_{90}$ after training with patterns at different levels of correlation and $fcorr_{90}$ with uncorrelated patterns ($fP$ = 0.0). It also shows this both when storing the prototypes themselves as well as when extracting them from distorted instances. Networks were of size $N$ = 400 with $K/H$ = 20. There were 20 runs per data point.

## Discussion

In this work we have compared quantitatively by means of extensive computer simulations seven different learning rules with regard to associative memory capabilities and focus on capacity to recall individual training patterns as well as the prototypes when training and testing with distorted instances. We additionally investigated resilience towards correlations, i.e. when training and testing with data from correlated sensors or feature detectors. Two different network architectures, non-modular and modular, were evaluated and compared and, generally, the difference between them was minimal. The non-modular networks stored somewhat fewer patterns but at the same time the information per pattern was higher. In the end, the per weight information capacity ($C$) turned out the same. From a neurobiological point of view, the modular network seems straightforward to realize with local lateral inhibition and divisive normalization provided by basket cells (Carandini et al., 1997; Lundqvist et al., 2010b), whereas the kWTA selection of maximally active units over a network comprising many hypercolumns seems more problematic to map to known neocortical architecture.

Importantly, associative memory performance varied considerably with learning rule used. The overall outcome was that the two Bayesian-Hebbian learning rules were superior with quite a large margin, to the others for individual pattern storage (Figures 5). The worst performing was the HEBB rule, which at the same time is the most popular in studies of associative memory. Our results compared quite well with previously published theoretical values and relations and also with the surprisingly scarce comparable simulation values found in literature, mainly derived using the original Hebbian learning rule. Unfortunately, a proper comparison is hampered by the lack of standardized capacity measures and variations in network architecture, retrieval procedure, and more.

For prototype extraction, the picture was somewhat different (Fig. 9). Firstly, the task was generally harder than individual pattern storage. In this task, the WILL rule failed entirely due to its binary weight matrix. However, the zip-net extension to this learning rule would remedy this problem (Knoblauch, 2010, 2016). The Bayesian-Hebbian rules again showed stable and significantly better performance than the others. Notably, we found no comparable published memory capacity values related to prototype extraction for any of the learning rules studied.

From the WILL rule example, we observe that, though using real valued weights does not increase pattern or weight information capacity *per se*, it does provide prototype extraction capability and robustness to input noise. This may be one reason why biological synapses are not binary but represent ca 5 bits of information (Bartol et al., 2015). This also suggests a possibility to significantly reduce the bit precision of connection weights and other plasticity state variables in neuromorphic hardware (Vogginger et al., 2015).

Regarding correlation sensitivity it can be noted that the PRCOV learning rule derived with the aim to reduce sensitivity to correlations indeed did that successfully. Especially for the prototype extraction case it was superior even to the Bayesian-Hebbian rules. However, another efficient and neurobiologically plausible way to handle correlated data, though not investigated here, is by building better internal representations, for instance, by creating one or more hidden layers of composite feature detecting units. As an example, this has been done with good results for BCPNN (Ravichandran et al., 2024, 2025).

A structural parameter of the networks fixed at $\sqrt{N}$ in our investigation was the sparsity of patterns, which also gives the size of hypercolumns for modular networks. This scaling scheme was suggested by Braitenberg (1978) on the basis of cortical architecture, though with the partitioning of cortex into hypercolumns[2] in mind. Here we apply this principle recursively, for partitioning hypercolumns into minicolumns. This partitioning scheme worked well in simulations of small to medium scale network models, altough we do not exclude other ways of partitioning. But the $\sqrt{N}$ scaling does not fit well when scaling to real cortex sizes because the estimated number of minicolumns per hypercolumn in cortex of higher mammals is limited, typically on the order of a hundred or so (Powell et al., 2024; Rockel et al., 1980). So for larger networks it is assumed that the expansion would occur in terms of increasing hypercolumn numbers alone. The full connectivity between minicolumns as in our simulated small networks is also unlikely to hold for scaled-up network models. Dilution of connectivity would obviously reduce the pattern storage capacity as measured in this study. In fact, the patchy connectivity described for neocortex of higher mammals (Koestinger et al., 2017) might result from this neocortical modularity and sparse axonal connectivity targeting specifically individual minicolumns.

[2] His columns correspond to our hypercolumns and his functional columns to our minicolumns.

## Summary amd Conclusions

We summarize performance over all benchmark tasks in Table 5. At the bottom of performance we find the HEBB and WILL learning rules. At the top we find the BCP and BOMs learning rules, which perform on par. Their pattern capacity is about 3x that of HEBB and close to 2x that of the runner up, PRCOV. Due to its high correlation resistance the latter reaches almost the same composite score as BCP and BOMs. In the end, task demands would determine how to weight the different task scores to select the best learning rule among the top performing.

The last row of Table 5 shows further that the modular architecture overall reached higher scores in these benchmarks than the non-modular one. Nevertheless, the information per weight is comparable since the modular network stores a higher number of patterns, each containing less information than the corresponding non-modular pattern.

A contribution to the superiority of the Bayesian-Hebbian rules could be their inclusion of intrinsic plasticity acting as a prior. This provides some extra tunable parameters, but adding such plasticity has good theoretical and neuroscientific support (Egorov et al., 2002). Another contribution might be that they were derived from a Bayes formalism for probabilistic inference operating in log space, thus combining evidence multiplicatively rather than additively.

| | Storage capacity scaling | | Weight info | Correlation resistance | | Learning rule scores | Storage capacity scaling | | Weight info | Correlation resistance | | Learning rule scores |
|---|---|---|---|---|---|---|---|---|---|---|---|---|
| | pattern | prototype | | pattern | prototype | | pattern | prototype | | pattern | prototype | |
| | | Non-modular | | | | | | Modular | | | | |
| HEBB | 6.2 | 5.4 | 7.1 | 4.7 | 7.1 | 6.1 | 6.6 | 6.0 | 6.8 | 3.7 | 7.4 | 6.1 |
| HOPF | 12.4 | 15.1 | 14.7 | 11.3 | 17.5 | 14.2 | 13.4 | 15.9 | 13.5 | 9.0 | 13.0 | 13.0 |
| COV | 12.5 | 15.2 | 14.8 | 15.7 | 19.7 | 15.6 | 13.5 | 15.9 | 13.5 | 14.2 | 13.9 | 14.2 |
| PRCOV | 12.9 | 16.5 | 7.5 | 26.1 | 25.8 | 17.8 | 13.7 | 16.7 | 7.0 | 25.0 | 38.0 | 20.1 |
| WILL | 13.8 | 0.0 | 14.0 | 4.0 | 0.0 | 6.4 | 13.3 | 0.0 | 14.4 | 3.3 | 0.0 | 6.2 |
| BCP | 21.0 | 24.2 | 21.2 | 17.7 | 12.3 | 19.3 | 19.9 | 23.0 | 22.5 | 21.9 | 14.1 | 20.3 |
| BOMs | 21.2 | 23.6 | 20.8 | 20.5 | 17.7 | 20.8 | 19.6 | 22.6 | 22.3 | 22.9 | 13.6 | 20.2 |
| Task scores | 44.0 | 41.6 | 50.5 | 42.70 | 40.5 | 41.6 | 56.0 | 58.4 | 49.5 | 57.3 | 59.5 | 58.4 |

**Table 5.** Grey values show relative learning rule performance over different tasks collected from the previously analyzed simulation results for a network size of N = 400 and H/K = 20. The average over all tasls of each learning rule is given in the framed orange middle and last columns. The bottom row shows averages over all taske for each learning rule, with non-modular and modular network scores summing to 100.

In conclusion, our learning rule benchmarking has provided important new information regarding what learning rules and neural mechanisms may be found in the brain as well as about learning and plasticity principles available for the design of future artificial neural systems. Notably, in our study, the popular HEBB learning rule came out with lowest overall performance. This has implications for considerations about learning and synaptic plasticity in the human brain as well as for engineering of future brain-like artificial intelligence artefacts.

## Methods

### *Iterative recall and stability*

During memory recall the network was iterated a maximum of fifteen steps. At lower memory load much fewer iterations were typically required to reach a stable attractor state. Due to the constraint of a fixed number of active units (K/H) conditions for convergence to a stable fix point was not always met. The fraction of unstable recalls increased with memory load and the instability was characterized by oscillations of cycle length two. Closer inspection revealed that in less than 1% of cases, the correct recall was among the oscillating states. The fraction of unstable retrievals was significantly higher for the BCP and BOMs learning rules but, even for those, at the loads where accuracy was tested ($P_{90}$), instability occurred in less than 1% of recalls and did not significantly affect reported recall performance.

### *Obtaining $P_{90}$ as measure of pattern capacity*

The left panel in Figure 2 exemplifies how the fraction of correct error-free recall ($f_{corr}$) from distorted patterns behaves when the number of random patterns in the training set is increased. To quantify the performance of the network in the pattern storage capacity task, we defined capacity as the maximum number of patterns that could be stored while maintaining an error-free recall of 90% of the (distorted) test patterns, here designated as $P_{90}$. This amounted to finding the number of patterns at which the curves in Figure 2 crossed the dotted 90 % line during decline. Since the process is probabilistic, a stochastic bisection method was used to estimate this crossing value. The same method was used to estimate storage capacity when training for the prototype extraction task.

The pseudo-code used for efficient estimate of the crossing with the 90% level is given below. *P0* was set to *N*. The estimated value and spread is mean and standard deviation of the last *P* of each run.

```
dir = 0 ; dirs = [] ; P = P0 ; d = 10% P0
while (len(dirs)<20 or abs(mean(dirs))>0.1) :
    corr = simulate(P,90) # Run the network with P patterns
    dir_old = dir
    dir = sign(corr - 90)
    P += dir * d
    if d>1 and dir*dir_old<0 :
       d = int(max(1,k*d + 0.5))
```

```
        elif d==1 :
            dirs := list of last 20 dir
 return P
```

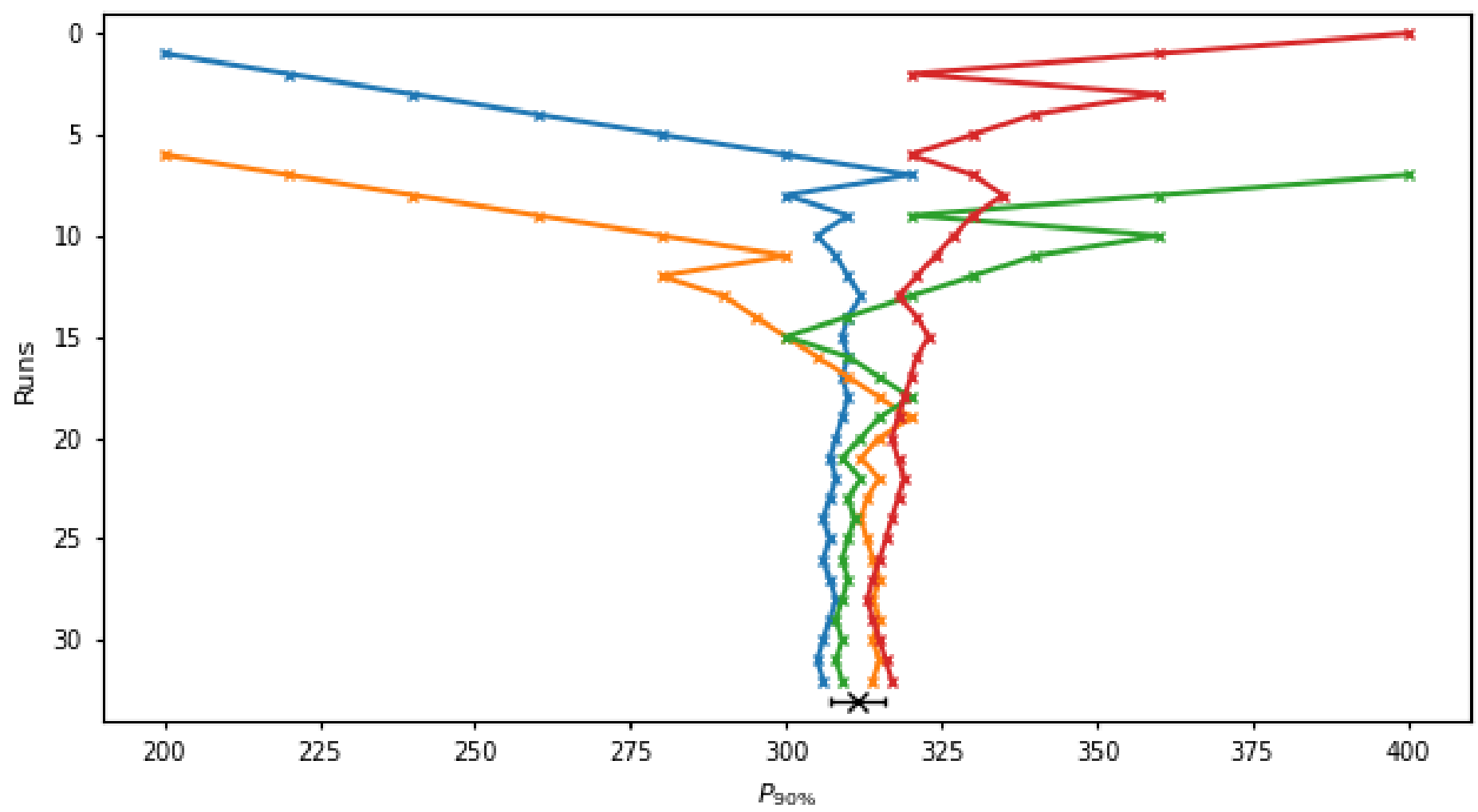

**Figure S1: Estimating storage capacity.** Output from the bisection method is shown with number of runs vs estimated $P_{90}$ for four runs of a BCP 16x16 network. P0 was 200 or 400 and the pattern distortion in each run resulted from resampling 2/16 hypercolumns. The resulting estimate mean was 312 and standard deviation 4.3.

### *Evaluation of weight information capacity*

The weight information capacity ($C = I/W$) was measured as the stored Shannon information ($I$) per trainable weight ($W$, free parameters). The total amount of retrievable information was calculated as mutual information between recalled and correct patterns. This was straigt-forward for the non-modular networks as patterns are just binary vectors. For the modular networks, the mutual information was calculated according to Appendix B. The number of trainable weights was calculated as $N(N-1)/2$ for non-modular networks and $N(N-H)/2$ for modular networks, except that for PRCOV which generates a non-symmetric weight matrix we did not divide by 2. It should be noted that the training pattern itself (0 input noise) can be exceptionally stable when inhibition gain is high as for the Bayesian-Hebbian learning rules.

This approach is similar to that by e.g Knoblauch and Palm (2020)), though here we applied it to iterative retrieval from a noisy cue (pattern correction) in a recurrent network rather than to pattern completion in a feed-forward network. From Figure 4 it is clear that number of stored patterns is independent of input noise at 25% and lower levels so we are in a robust-retrieval regime where the pattern capacity is independent of input noise and instead determined by the amount of cross-talk between patterns and stability of the stored attractor. Our $I/W$ value is therefore comparable to the bits per weight values often given in literature mostly derived from theoretical calculations.

### *Evaluation of pattern correlation resistance*

We measured on 20x20 sized non-modular and modular network trained with single patterns and for prototype extraction, for an initial assessment of correlation resistance. The slope ($|k|$) of a line through (0,0] fit to the four first data points was used as a measure of sensitivity, and an index of "correlation resistance" was calculated as -1/$k$.

### *Calculating scores for the summary Table 5*

The values in the gray areas of the table gives relative performance values within each task, so they sum to 100 column wise. The values in the corresponding pink columns show the mean for each learning rule over all tasks, separate for non-modular and modular networks. The bottom row shows relative performance of non-modular and modular networks, calculated from the summed absolute performance in each task and scores for non-modular and modular networks sum to 100 for each task.

## Author contributions: CRediT

**Anders Lansner:** Conceptualization, Methodology, Software, Validation, Writing original draft, Visualization, Project administration; **Andreas Knoblauch:** Software, Formal analysis, Writing, reviewing & editing; **Naresh Ravichandran:** Software, Writing, reviewing & editing; **Pawel Herman:** Resources, Writing, reviewing & editing, Supervision, Funding acquisition

## Declaration of competing interest

The authors declare no conflict of interest.

## Acknowledgements

Funding for the work was received from the Swedish e-Science Research Centre (SeRC), Digital Futures, Swedish Research Council (VR2018-05360 and VR2016-05871), the European Commission Directorate-General for Communication Networks, Content and Technology grant no. 101135809 (EXTRA-BRAIN) and Indo-Swedish Joint Network 2018 grant No. 2018–07079,.

Appendix A – The Bayes Optimal Memory learning rule (BOM, BOMs)

The original BOM learning rule minimizes output noise and maximizes storage capacity by activating neurons based on Bayesian maximum likelihood decisions [Knoblauch, 2011, 2024]. In the auto-associative case, the task is to store $M$ activity patterns $u^\mu$, where $\mu = 1, \dots, M$. Here the $u^\mu \in \{0,1\}^N$ are binary vectors of size $N$. In the original formulation, associations are stored using first order (neural) and second order (synaptic) counter variables

$$M_u(j) := \left|\{\mu: u_j^\mu = u\}\right|$$

$$M_{uv}(ij) := \left|\{\mu: u_i^\mu = u, u_j^\mu = v\}\right|$$

where $u, v \in \{0,1\}$ and $i, j \in \{1,2,\dots,N\}$. Note $M_1(i) = C_i$, and $M_{11}(ij) = C_{ij}$ for the counter variables of Table 1. Note that it is sufficient to store only $M$, $M_1$, and $M_{11}$, as all other counters can be reconstructed from

$$M_0(j) = M - M_1(j) \quad \text{and} \quad M_{uv}(ij) = M_v(j) - M_{(1-u)v}(ij)\ .$$

For retrieval of stored memories the general BOM model assumes noisy query patterns $\tilde{u}$ resembling one of the stored memories $u^\mu$, where

$$q_{uv|w}(ij) := \mathrm{pr}\left[\tilde{u}_i = v | u_i^\mu = u, u_j^\mu = w\right]$$

defines "noise" as the probability of a bit switching from $u$ to $v$, conditioned on the postsynaptic activity $w$ of the original memory pattern. With this, biases $b_j$ and synaptic weights $w_{ij}$ (from neuron $i$ to neuron $j$) of BOM write

$$b_j := (n-1)\log\frac{M_0}{M_1} + \sum_{i=1}^{n} \log\frac{M_{01}\left(1-q_{01|1}\right) + M_{11}q_{10|1}}{M_{00}\left(1-q_{01|0}\right) + M_{10}q_{10|0}}$$

$$w_{ij} := \log\frac{\left(M_{11}\left(1-q_{10|1}\right) + M_{01}q_{01|1}\right)\cdot\left(M_{00}\left(1-q_{01|0}\right) + M_{10}q_{10|0}\right)}{\left(M_{10}\left(1-q_{10|0}\right) + M_{00}q_{01|0}\right)\cdot\left(M_{01}\left(1-q_{01|1}\right) + M_{11}q_{10|1}\right)}$$

where $n$ is the number of recurrent connections neuron $j$ receives (e.g., n=N-1 for networks without autapses). Note also that we have skipped indexes $i, j$ of counter variables and error probabilities for brevity, i.e., $M_u := M_u(j), M_{uv} := M_{uv}(ij), q_{uv|w} := q_{uv|w}(ij)$. Then the optimal decision to activate a component of the retrieval output $\hat{u}$ is

$$\hat{u}_j = 1 \quad \text{if} \quad x_j := b_j + \sum_{i=1}^{n} w_{ij}\tilde{u}_i \geq 0$$

where the "dendritic potential" $x_j$ corresponds to the log-odds-ratio $\log\left(\mathrm{pr}\left[u_j^\mu = 1 \middle| \tilde{u}\right] / \mathrm{pr}\left[u_j^\mu = 0 | \tilde{u}\right]\right)$.

For the experiments of this study we proposed the BOMs model, which assumes zero noise estimates $p_{uv|w}(ij) := 0$ because we focus on high quality iterative retrieval, where initially noisy inputs typically improve over iteration steps to the orignal memory patterns, without any errors. A more optimal implementation could adapt the estimates $p_{uv|w}(ij)$ over iteration steps, but this would be computationally more expensive and more difficult to justify biologically. Then the BOMs equations for bias and weights simplify to

$$b_j = (n-1)\log\frac{M_0}{M_1} + \sum_{i=1}^{n}\log\frac{M_{01}}{M_{00}} = (n-1)\log\frac{1-p_j}{p_j} + \sum_{i=1}^{n}\log\frac{p_j - p_{ij}}{1-p_i-p_j+p_{ij}}$$

$$w_{ij} = \log\frac{M_{11}\cdot M_{00}}{M_{10}\cdot M_{01}} = \log\frac{p_{ij}\cdot(1-p_i-p_j+p_{ij})}{(p_i-p_{ij})\cdot(p_j-p_{ij})} \quad (= w_{ji} \text{ for auto} - \text{association}).$$

using the notation of Table 2, where we extended all fractions by $M$ after replacing

$$M_0(i) = M - M_1(i)$$
$$M_{01}(i,j) = M_1(j) - M_{11}(i,j)$$
$$M_{00}(i,j) = M_0(i) - M_{01}(i,j) = M - M_1(i) - M_1(j) + M_{11}(i,j)$$
$$M_{10}(i,j) = M_0(j) - M_{00}(i,j) = M_1(i) - M_{11}(i,j).$$

Estimating the stabilization parameter $\epsilon$

To avoid division by 0 and log(0) we bounded nominators and denominators by a small numerical stabilization parameter $\epsilon > 0$,

$$b_j \coloneqq (n-1)\log\frac{\max(1-p_j,\epsilon)}{\max(p_j,\epsilon)} + \sum_{i=1}^{n}\log\frac{\max(p_j-p_{ij},\epsilon)}{\max(1-p_i-p_j+p_{ij},\epsilon)}$$

$$w_{ij} \coloneqq \log\frac{\max\big(p_{ij}(1-p_i-p_j+p_{ij}),\epsilon\big)}{\max\Big((p_i-p_{ij})(p_j-p_{ij}),\epsilon\Big)}.$$

Similarly, for BCP we use $b_j \coloneqq \log\max(p_j,\epsilon)$, $w_{ij} \coloneqq \log\frac{\max(p_{ij},\epsilon)}{\max(p_i p_j,\epsilon)}$ and for PRCOV $w_{ij} \coloneqq \frac{p_{ij}-p_i p_j}{\max(p_i,\epsilon)}$ (cf., Table 2).

Assuming random patterns with independent components $u_i^\mu$ and $a \coloneqq pr[u_i^\mu = 1]$, the counter variables have binomial distributions, e.g., $M_{11} \sim B(M, a^2)$. It is easy to see that for large networks ($n \to \infty$), sparse activity ($a \to 0$), and high memory load close to the capacity limit $M \sim \frac{n}{-a(1-a)\log a}$ (e.g., see Knoblauch, 2011, eq. 2.37), the expected numbers $E_u, E_{uv}$ of counter variables $M_u(i), M_{uv}(ij)$ being zero quickly become zero, except for the coincidence counters $M_{11}(ij)$ having diverging $E_{11} \to \infty$. This means that, in practice, most cases of lower bound clipping is due to $M_{11}(ij) = 0$ and, consequently, we may identify the stability parameter $\epsilon$ or rather $\frac{\epsilon}{M}$ with the relevant error term $M_{01}q_{01|1}$ in the general equation of $w_{ij}$. This leads to the following reasonable estimate

$$\epsilon \coloneqq \frac{-a\cdot\log(p_{corr})}{n}$$

where $p_{corr}$ is the tolerated (minimal) fraction of correct retrieval outputs.

We used the same $\epsilon$ for BCP given in Table 2, since it can be understood as a simplification of BOMs using the approximations $1 - p_i - p_{ij} \approx 1$, $p_i - p_{ij} \approx p_i$ and $p_j - p_{ij} \approx p_j$ valid for sparse activity where $p_i, p_j << 1$ and $p_{ij} << p_i, p_j$. A different way of setting $\epsilon$ was previously used for BCP with good results (Martinez Mayorquin, 2022).

Appendix B – On mutual information for modular networks with WTA activation

For evaluation of storage capacity in terms of Shannon information we generally follow Knoblauch/Palm (2020). For modular networks storing "block patterns" consisting of K blocks (=hypercolumns or modules) of size M, we can assume here, due to strict winner-takes-all retrieval, that each block has exactly one active neuron. If two or more neurons within a "block" receive the same amount of synaptic input, only one neuron will get activated at random.

Similar to Knoblauch and Palm (2020, Appendix A), we need to compute the transinformation between an original Block $X \in \{0,1\}^M$ and a retrieved block $Y$ of a memory pattern. For random block patterns (with uniform distribution of the active unit within each block), the Shannon information (in bits) contained in a single block is obviously

$$I(X) = \mathrm{ld}(M)$$

Now let $p_{\mathrm{err}}$ be the probability that a retrieved block Y is not correct. For correct blocks Y the conditional surprise of $X$ given $Y$ is simply $S(X|Y) = -\mathrm{ld}(P[X = Y]) = -\mathrm{ld}(1 - p_{\mathrm{err}})$, whereas for an incorrect blocks Y we have $S(X|Y) = -\mathrm{ld}\big(P(X|Y)\big) = -\mathrm{ld}(p_{\mathrm{err}} \cdot \frac{1}{M-1})$, as excluding the incorrect unit of Y, there are still $M - 1$ units to choose from the correct unit. Here we assume that each of these units has an equal chance of being the correct one (which is plausible if we consider many blocks from many different networks). Now the corresponding conditional Shannon information $I(X|Y)$ of X given Y is the expected surprise $S(X|Y)$ averaged over all cases,

$$\begin{aligned} I(X|Y) = E[S(X|Y)] &= -(1 - p_{\mathrm{err}})\mathrm{ld}(1 - p_{\mathrm{err}}) - p_{\mathrm{err}}\mathrm{ld}\left(p_{\mathrm{err}} \cdot \frac{1}{M-1}\right) \\ &= I(p_{\mathrm{err}}) + p_{\mathrm{err}}\mathrm{ld}(M-1) \end{aligned}$$

where $I(p) = -p\mathrm{ld}(p) - (1 - p)\mathrm{ld}(1 - p)$ is the Shannon information of a binary random variable $Z \in \{0,1\}$ that has probability $p = P[Z = 1]$ of being one. Thus, the transinformation $T(X;Y)$ between X and Y computes (similar to Knoblauch/Palm, 2020, eq.(A.8))

$$T(X;Y) = I(X) - I(X|Y) = \mathrm{ld}(M) - I(p_{\mathrm{err}}) - p_{\mathrm{err}}\mathrm{ld}(M-1)$$

Thus, for a network storing P memory patterns, the stored Shannon-(trans-)information per synaptic weight (corresponding to the „mapping capacity" of Knoblauch/Palm (2020; eq. 2.7) is

$$C_{\mathrm{v}}(p_{\mathrm{err}}) = \frac{P \cdot K \cdot T(X;Y)}{\#\mathrm{synapses}} = \frac{P \cdot K \cdot (\mathrm{ld}(M) - I(p_{\mathrm{err}}) - p_{\mathrm{err}}\mathrm{ld}(M-1))}{\#\mathrm{synapses}}$$

Correspondingly, the „completion capacity" corresponding to Knoblauch/Palm (2020; eq. 2.8), that is, the difference between the mapping capacities of outputs and inputs, is

$$\begin{aligned} C_{\mathrm{u}} &= C_{\mathrm{v}}\big(p_{\mathrm{err,out}}\big) - C_{\mathrm{v}}\big(p_{\mathrm{err,in}}\big) \\ &= \frac{P \cdot K \cdot [I\big(p_{\mathrm{err,in}}\big) - I\big(p_{\mathrm{err,out}}\big) + \big(p_{\mathrm{err,in}} - p_{\mathrm{err,out}}\big) \cdot \mathrm{ld}(M-1)]}{\#\mathrm{synapses}} \end{aligned}$$

where $p_{\mathrm{err,in}}$ and $p_{\mathrm{err,out}}$ are block error probabilities of input patterns and retrieval outputs, respectively. Thus, to compute information-based capacities, we only require block error probability $p_{\mathrm{err}}$ for input- and output patterns. Assuming independent block errors, the fraction of correct retrievals is $p_{\mathrm{corr}} = (1 - p_{\mathrm{err}})^K$, and we may use

$$p_{\mathrm{err}} = 1 - p_{\mathrm{corr}}^{1/K}$$

as, in some experiments, we have evaluated only $p_{\mathrm{corr}}$.

Supplementary figure S1

Scaling of $C$ with $N$ depending on learning rule and input noise level, showing that the value typically converges with incresing network size.

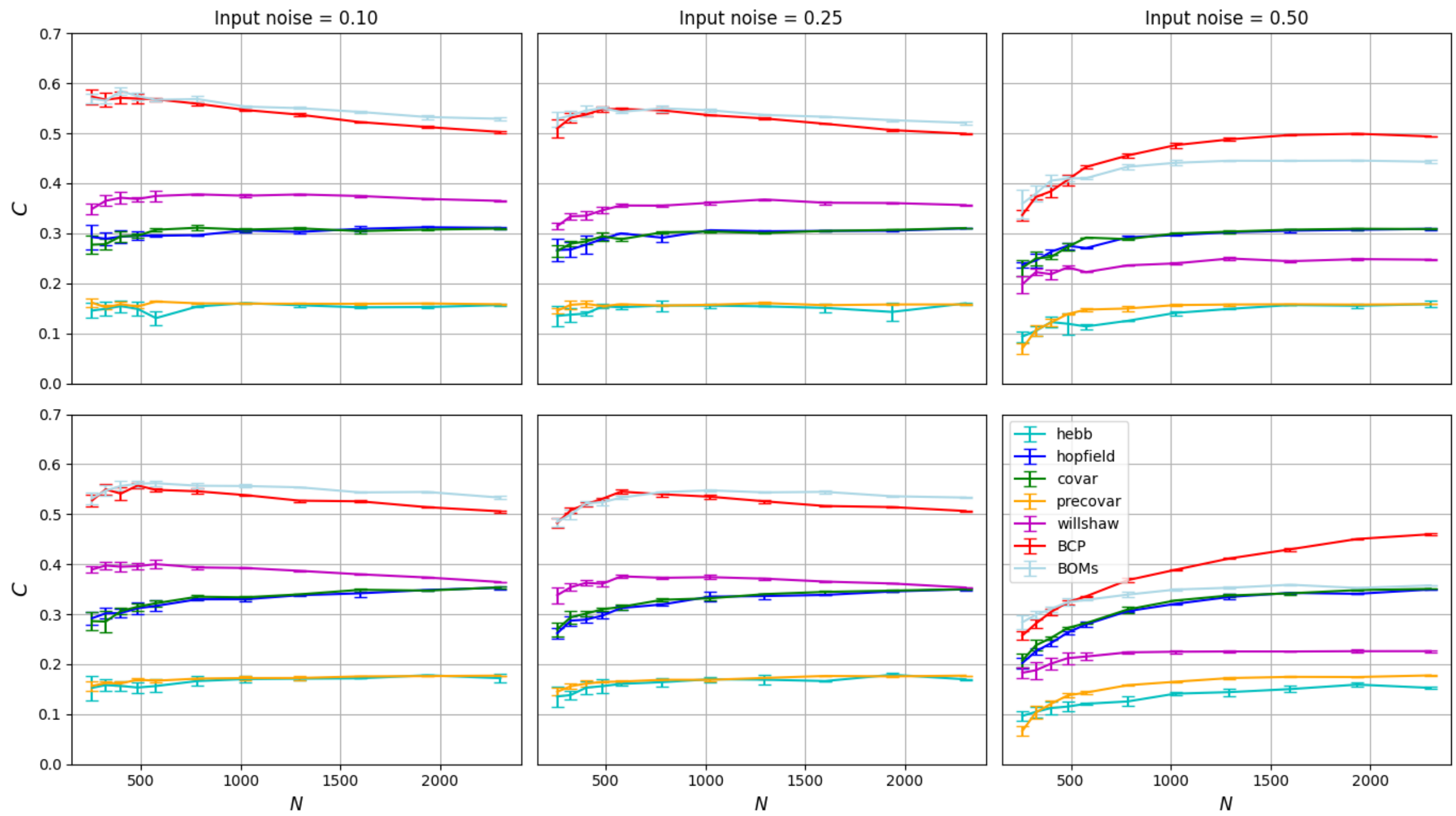